\documentclass[letterpaper]{article}

\usepackage[numbers]{natbib}
\usepackage{alifeconf}
\usepackage{url,hyperref,cleveref}
\usepackage{booktabs}

\usepackage{lipsum}

\newcommand\blfootnote[1]{%
  \begingroup
  \renewcommand\thefootnote{}\footnote{#1}%
  \addtocounter{footnote}{-1}%
  \endgroup
}

\usepackage{xcolor}

\newif\ifanonymous
\anonymousfalse   

\title{When Does Structure Matter in Continual Learning? Dimensionality Controls When Modularity Shapes Representational Geometry}

\author{
\ifanonymous
    Anonymous Authors \\
    Anonymous Institution \\
    \texttt{anonymous@anonymous.com}
\else
    Kathrin Korte$^{1}$,
    Joachim Winter Pedersen$^{1}$,
    Eleni Nisioti$^{1}$, \and
    Sebastian Risi$^{1}$ \\
    \mbox{}\\
    $^1$IT University of Copenhagen, Denmark \\
    kort@itu.dk
\fi
}

\ifanonymous
\hypersetup{
    pdfauthor={Anonymous},
    pdftitle={Anonymous Submission}
}
\fi

\begin{document}

\maketitle

\begin{abstract}
 
        
        To preserve previously learned representations, continual learning systems must strike a balance between plasticity, the ability to acquire new knowledge, and stability.   This stability-plasticity dilemma affects how representations can be reused across tasks: shared structure enables transfer when tasks are similar but may also induce interference when new learning disrupts existing representations. However, it remains unclear when and why structural separation influences this trade-off. In this study, we examine how network architecture, task similarity and representational dimensionality jointly shape learning in a sequential task paradigm inspired by transfer–interference studies. We compare a task-partitioned modular recurrent network with a single-module baseline by systematically varying task similarity (low, medium, high) and the scale of weight initialization, which induces different learning regimes that we empirically characterize through the effective dimensionality of the learned representations. We find that architecture has minimal impact in high-dimensional regimes where representations are sufficiently unconstrained to accommodate multiple tasks without strong interference. In contrast, in lower-dimensional (rich) regimes, architectural separation is decisive: modular networks exhibit graded alignment of task-specific subspaces with overlap for similar tasks, partial orthogonalization for moderately dissimilar tasks and stronger separation for dissimilar tasks. This graded geometry is absent in the single network baseline. Our findings suggest that representational dimensionality acts as a key organizing variable governing when structural separation becomes functionally relevant, and highlight adaptive geometry as a central principle for designing continual learning systems.

\end{abstract}

Submission type: \textbf{Full Paper}\\


Data/Code available at: \href{https://github.com/kat-ko/Dimensionality-Controls-When-Modularity-Shapes-Representational-Geometry}{github.com/Dimensionality-Controls-When-Modularity-Shapes-Representational-Geometry}

\ifanonymous\else
\blfootnote{\textcopyright  2026 [Kathrin Korte, Eleni Nisioti, Joachim Winter Pedersen, Sebastian Risi]. Published under a Creative Commons Attribution 4.0 International (CC BY 4.0) license.}
\fi

\section{Introduction}

        
        Continual learning is a central challenge for artificial and biological intelligence: an adaptive system must learn from a stream of experiences, integrate new information, and remain effective on previously encountered tasks \cite{ramasesh2020anatomy}. This is difficult because sequential learning creates a fundamental tension between plasticity, to acquire new knowledge, and stability, to preserve previously learned representations \cite{holton2025humans}. In practice, this tension is reflected in how existing representations are reused: shared structure can support transfer across tasks, but may also lead to interference when new learning disrupts prior knowledge \cite{lee2021continual, liu2023connectivity, mathis2025leveraging}. In this sense, continual learning is not only a problem of memory retention, but also a problem of how learned structure should be reorganized in response to changing task demands. Learning depends not only on what is learned, but also on how experience is structured and represented over time \cite{menghi2025impact}. 
        
        A key factor in this trade-off is task similarity \cite{menghi2025effects}. When successive tasks are highly related, reusing existing representations is often advantageous because it enables transfer and faster adaptation. When tasks are dissimilar, however, the same reuse may become harmful because it increases interference \cite{hiratani2024disentangling}. This creates a non-trivial decision problem: should a learner continue to adapt an existing representation, or allocate new representational resources and keep tasks more separate? In the language of psychology, this resembles the tension between lumping and splitting: learners may either consolidate related experiences into a shared structure or carve out distinct representations when task demands diverge. The optimal strategy depends on the current task, the similarity to prior experience, and the learner's expectations about future tasks \cite{holton2025humans}.
        
        Several mechanisms have been proposed to mitigate interference in sequential learning. Replay-based approaches and synaptic consolidation methods, such as Elastic Weight Consolidation (EWC), seek to protect previously acquired knowledge by constraining parameter updates or revisiting old experiences. \cite{kirkpatrick2017overcoming, rolnick2019experience} Architectural approaches aim to reduce interference through structural separation, for example, by assigning different tasks to distinct modules or pathways \cite{ellefsen2015neural}. Modularity is attractive because it can isolate task-specific computations and thereby reduce catastrophic forgetting. Yet this comes at a cost: in the extreme case of complete structural separation, interference may be eliminated, but so may transfer \cite{holton2025humans}. 
        
        This raises a deeper question: how can a learner preserve useful prior knowledge while still avoiding harmful overwrite? Put differently, what determines whether task representations should overlap, partially align, or separate? One promising answer is that the relevant variable is not only architecture, but the representational dimensionality and representational geometry that the network can realize. Here, \emph{representational geometry} refers to the similarity and distance structure of hidden-state activity patterns, and \emph{representational dimensionality} refers to the number of independent directions those patterns occupy \cite{kriegeskorte2013representational}. Recent work has shown that representations in a particular brain region can contain both low- and high-dimensional components \cite{johnston2024modular}, and that rich versus lazy learning regimes can produce markedly different representational structures \cite{yu2025dimensionality}.
        During lazy learning, the input dimensionality is expanded by random projections to the network's hidden layer, whereas in rich learning, hidden units acquire structured representations that privilege relevant over irrelevant features \cite{flesch2021rich}.
        Modular organization in multi-task learning has been shown to emerge only under certain input and task conditions \cite{Johnston2024.09.30.615925, yue2017brain}. Results by Johnston et al. suggest that when inputs are low-dimensional and structured, an implicit modular organization can evolve, whereas high-dimensional unstructured inputs may yield more diffuse or implicit forms of organization \cite{Johnston2024.09.30.615925}. These findings point to a broader possibility: structural bias may only influence continual learning behavior when the representational space is sufficiently constrained for geometry to matter.
        
\begin{figure*}[!t]
            \centering
            \includegraphics[width=\textwidth]{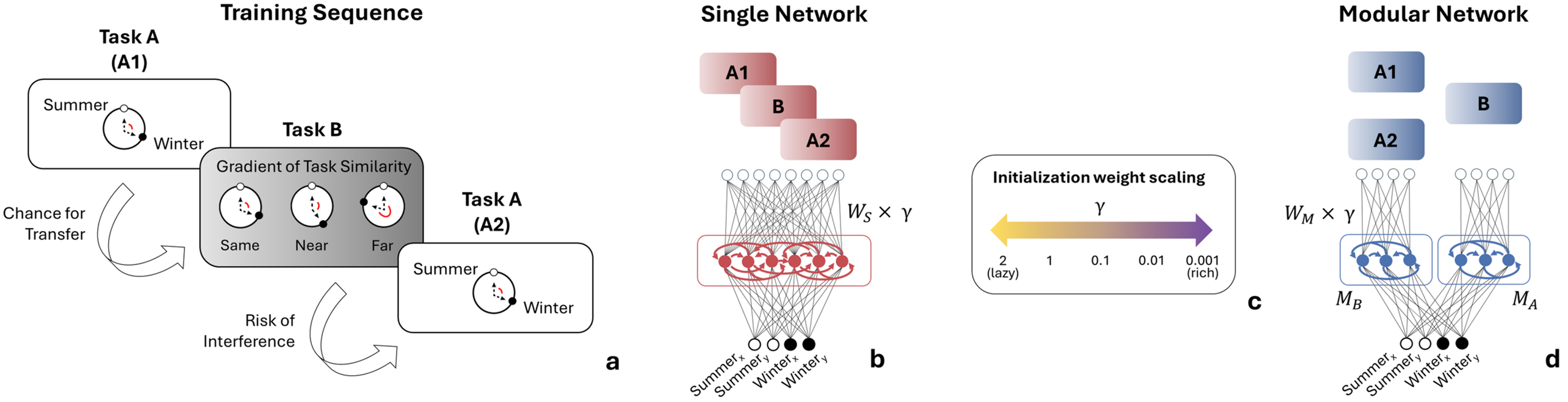}
            \caption{\textbf{Overview of the continual-learning setup, architectures, and representational regimes.} \textbf{(a)} Sequential training protocol. Networks are trained on task A in phase $A1$, then on task B in phase $B$, and finally retested on task A in phase $A2$. Task B is instantiated in three similarity conditions relative to task A: same, near, and far. The schematic highlights two competing pressures: the opportunity for transfer when tasks are similar and the risk of interference when learning task B alters representations used for task A. \textbf{(b)} Single-network baseline. All inputs are processed through one recurrent population, and predictions are produced from a shared readout. \textbf{(c)} Weight scaling. After initialization, all trainable weights (Single: $W_S$, Modular: $W_M$ ) are rescaled by a factor \(\gamma\), spanning the lazy, high-dimensional regime (\(\gamma\) large) to the rich, low-dimensional regime (\(\gamma\) small). \textbf{(d)} Task-partitioned modular network. Inputs are routed to task-specific recurrent modules and module outputs are combined by a shared readout. This architecture imposes structural separation and allows us to test when such separation alters learning dynamics and representational geometry.}
            \label{fig:overview}
        \end{figure*}

        In this paper, we examine this hypothesis using a sequential transfer-interference paradigm inspired by prior work on task similarity and representational separation (Figure \ref{fig:overview}a) \cite{holton2025humans}. We compare a task-partitioned modular recurrent network (Figure \ref{fig:overview}d) with a single-module baseline (Figure \ref{fig:overview}b), and systematically vary both task similarity (same, near, far) and initialization scale (Figure \ref{fig:overview}c). As in \citet{holton2025humans}, we treat initialization scale as a control variable that changes the effective dimensionality of learned representations, and we ask whether architecturally induced structural bias only becomes meaningful in continual learning once the representational regime becomes sufficiently constrained. To address this, we analyze behavioral outcomes such as accuracy, transfer, and interference, as well as the geometry of hidden-state representations, using effective dimensionality, principal angles, and qualitative 3D PCA visualizations of task-specific trajectories.
        Our main result is that structural separation is not uniformly beneficial. Instead, the effect depends on the representational regime. In high-dimensional settings, modular and single networks behave similarly and show limited differences in representational geometry or continual learning performance. In lower-dimensional settings, however, modularity becomes functionally meaningful: it supports a graded representational organization in which similar tasks remain more aligned, moderately similar tasks occupy intermediate subspaces, and dissimilar tasks separate more strongly. This suggests that continual learning is best understood as a problem of adaptive representational allocation, where the key question is not whether representations should be shared or separated in general, but when and how the network should reorganize them. From this perspective, representational dimensionality becomes a control variable that determines when modularity matters, and task similarity determines the form that separation should take.
        
        Our study results in two main insights. First, it shows that modularity does not automatically solve continual learning; its effect depends on the geometry of the representation space. Second, it suggests that successful continual learning requires a similarity-dependent balance between overlap and separation, rather than maximal isolation. More broadly, this work connects continual learning to a geometric view of adaptation in complex systems: to remain flexible and stable over time, a learner must regulate not only its parameters, but also the dimensionality and organization of the representational space in which new tasks are encoded.

\section{Background and Related Work}

\subsubsection{Continual Learning, Task Similarity, and Representational Geometry}

Continual learning studies how a learner can acquire new tasks without forgetting previously acquired knowledge. The central difficulty is the stability-plasticity trade-off: preserving earlier task performance requires limiting updates to established representations, whereas adaptation to new tasks often benefits from reusing and reshaping those same representations \cite{lee2021continual,holton2025humans}. This tension has motivated regularization-based methods, rehearsal, replay, and architectural separation \cite{wickramasinghe2023continual}. EWC \cite{kirkpatrick2017overcoming} mitigates catastrophic forgetting by selectively constraining changes to parameters that are important for previously learned tasks; it approximates the posterior over Task A parameters with a diagonal Gaussian centered at $\theta^{*}_{A}$ and uses the Fisher information to weight the elastic constraint according to parameter importance. Replay, by contrast, interleaves stored examples from prior tasks with training data from the current task, reinforcing previously learned representations and preserving Task A performance without requiring architectural modifications \cite{robins1995catastrophic,rolnick2019experience,van2020brain}. This corresponds to rehearsal-based approaches.

A key determinant of whether reuse helps or harms is task similarity. When successive tasks are highly related, reusing existing representations can support forward transfer; when tasks are dissimilar, the same reuse can cause interference and catastrophic forgetting \cite{lee2021continual,menghi2025impact,holton2025humans, wakhloo2026neural}. This creates a non-trivial decision problem: whether to continue adapting an existing representation or to allocate more separated representational resources for the new task. Representational geometry provides a natural language for this problem by asking whether different tasks are encoded in overlapping, partially aligned, or approximately orthogonal subspaces \cite{flesch2022representation}. In both neuroscience and machine learning, this perspective shifts attention from task performance and control processes alone to the internal format of the representations that support behavior \cite{flesch2022representation}. Rich and lazy learning regimes induce different representational geometries and trade off learning speed against robustness: rich regimes support feature learning and more structured, often low-dimensional internal codes, whereas lazy regimes preserve representations closer to their initialization and often yield higher-dimensional but less specialized solutions \cite{flesch2022representation, seguin2022network}. 

Although low-dimensional activity can be a signature of system consolidation at the end of learning, high initial dimensionality may also be useful when probing new neural patterns in search of optimal control \cite{gurnani2023signatures}. Holton et al.~\cite{holton2025humans} show in a changing seasons-style sequential learning task that representational overlap and transfer are systematically modulated by task similarity. Their results indicate that low-dimensional representations in a rich regime can support orthogonalization when tasks are dissimilar, while still allowing reuse and overlap when tasks are similar. In intermediate similarity conditions, the learned geometry can remain partially aligned, yielding transfer while also increasing vulnerability to interference upon retest. Menghi et al.~\cite{menghi2025impact} likewise show that the interaction between task similarity and training regime shapes whether learners exhibit transfer or interference, indicating that the temporal organization of experience matters in addition to task content.

\subsubsection{Dimensionality, Modularity, and Architectural Constraints}

These questions are not limited to artificial systems. Population activity in biological circuits often concentrates on low-dimensional manifolds that capture task-relevant latent variables and dynamics, but the relationship between representational dimensionality and performance is heterogeneous across brain regions \cite{yu2025dimensionality}. In some settings, compressed representations support stronger performance, whereas in others, higher-dimensional representations preserve important information and better support behavior \cite{yu2025dimensionality}. More generally, there is no simple monotonic relationship between dimensionality and success: high- and low-dimensional representations can support different forms of generalization, compression, and robustness, and the benefit of compression depends on task structure and on the constraints imposed by architecture and optimization dynamics \cite{flesch2022representation,yu2025dimensionality}. For continual learning, this implies that the usefulness of architectural separation may depend on whether the representational space is sufficiently constrained for geometry to become a binding factor.

A complementary line of work studies modularity as an architectural prior \cite{ellefsen2015neural, mathis2025leveraging, achterberg2023spatially}. Modular architectures mitigate interference by allocating distinct subsets of network parameters to different tasks, thereby reducing overlap in representational resources and decreasing catastrophic interference, but they also limit the possibility of transfer through shared computation. This makes modularity a useful structural bias for studying the balance between reuse and separation in continual learning \cite{salatiello2026modularitybedrocknaturalartificial}. Prior work has shown that modular organization can arise in both biological and artificial systems and can support adaptive behavior, especially when tasks compete for limited representational resources \cite{clune2013evolutionary,yue2017brain,ellefsen2015neural}. In cognitive neuroscience, modularity has been linked to task complexity and performance, and theoretical work has argued that specialized subpopulations can emerge naturally under constraints on connectivity and physical arrangement \cite{yue2017brain, gu2024emergence}. Béna et al.~\cite{bena2025dynamics} show that specialization within modular recurrent systems can change dynamically depending on information availability and resource constraints, even when coordination is mediated only through a shared readout. When information is globally accessible, modules may de-specialize; when access is constrained, division of labor becomes more pronounced. This suggests that structure and function are tightly coupled and that specialization is not a fixed property of a module but an emergent response to structural constraints on information flow.

The effect of architecture on representational geometry may itself depend on the dimensionality of the available representation space. Johnston et al.~\cite{Johnston2024.09.30.615925} argue that high-dimensional input representations can allow even complex tasks to be solved by a simple readout, without requiring explicit modularity. Their work also shows that explicit modularity emerges most clearly under low-dimensional input representations, whereas high-dimensional unstructured inputs yield representations that are either unstructured or only implicitly modular, depending on the output structure \cite{Johnston2024.09.30.615925}. This suggests that structural separation may only become functionally meaningful when representational capacity is sufficiently constrained. Similarly, Lu et al.~\cite{lu2025rethinking} argue that the stability-plasticity trade-off is not only a learning-rule problem, but also an architectural one: under fixed parameter budgets, different network structures can yield different balances between stability and plasticity. More broadly, specialization can be understood as an emergent consequence of how task structure is mapped into a constrained representational space. Task similarity then determines whether reuse is beneficial, whether partial overlap is sufficient, or whether a more complete separation is needed to avoid interference.

Taken together, the literature suggests that continual learning is shaped jointly by task similarity, representational geometry, dimensionality, and architectural structure. Some work emphasizes learning regimes and hidden-state geometry \cite{flesch2022representation}, some focuses on similarity-driven transfer and interference \cite{holton2025humans,menghi2025impact}, and others examine modularity or dimensionality as architectural and representational constraints \cite{yu2025dimensionality,Johnston2024.09.30.615925,bena2025dynamics,lu2025rethinking}. The present study builds on these strands by asking how a task-partitioned modular recurrent architecture, compared with a single-network baseline, organizes representational geometry under graded task similarity and varying representational regimes. In doing so, it connects the literature on continual learning and interference to the broader question of when structural separation becomes functionally meaningful.

\section{Methods}
        \subsection{Experimental Paradigm}
        
        We studied continual learning in a sequential transfer--interference task originally introduced by Holton et al.~\cite{holton2025humans}. We use this task only as a controlled experimental paradigm for studying continual learning in neural networks, not as a model of human behavior. The task consists of three phases: \(A1 \rightarrow B \rightarrow A2\) (Figure~\ref{fig:overview}a). In phase \(A1\), the network learns task A. In phase \(B\), it learns task B on a new set of stimuli. In phase \(A2\), task A is revisited to quantify retention and interference after exposure to task B (Figure \ref{fig:overview}a). Each task requires mapping six discrete plant cues to positions on a circular dial. For each plant, locations in the two seasonal contexts, \emph{summer} and \emph{winter}, are related by a fixed angular offset, referred to as the \emph{task rule}. Knowing a plant's location in one season and the task rule determines its location in the other season. Across schedules, the numerical value of the rule is randomized. Task B uses novel stimuli but the same formal structure as task A. The \emph{Same}, \emph{Near}, and \emph{Far} conditions manipulate how the task B rule relates to the task A rule: the rules are identical in the Same condition, shifted by a small angle in the Near condition, and shifted by a large angle in the Far condition. All other design features are held fixed across these conditions~\cite{holton2025humans}. Recurrent networks were trained on trial sequences derived from the original experiment. These sequences matched the original task structure in trial order, stimulus identity, probed season, feedback availability, and task rule. The networks were trained independently and were not fit to human motor responses. Each stimulus was encoded as a one-hot input vector. The network produced a four-dimensional output consisting of two cosine-sine pairs: the first pair encoded the summer location (\texttt{feature\_idx}=0), and the second pair encoded the winter location (\texttt{feature\_idx}=1). On each trial, mean squared error (MSE) was calculated only for the output pair corresponding to the probed season, while the other pair did not contribute to the loss for that step. Angular accuracy was computed by reconstructing the predicted angle from the supervised cosine-sine pair and comparing it to the target angle on the circle.
        
        Training was organized in three phases: \(A1\) (task A acquisition), \(B\) (task B acquisition on new stimuli), and \(A2\) (retest on task A). Each phase followed the same trial-wise alternation between summer and winter probes as in the original design. Following Holton et al., transfer was quantified using winter-trial accuracy: the change from the end of task A to the beginning of task B, operationalized in our analysis pipeline as the mean over the last six winter trials in \(A1\) versus the first six winter trials in \(B\). Interference at task A retest was quantified from winter responses in phase \(A2\) using the mixture-based measure described in the Analysis section, analogous to Holton et al.'s probability-of-updating-to-the-task-B-rule treatment. Where we report generalization within task A, this refers to winter performance on held-out stimuli following the paper-faithful testing protocol when applicable.
        
        \subsection{Architectures}
        
        We compared two recurrent architectures: a single-network baseline and a task-partitioned modular network. The single-network baseline processes all inputs through one recurrent population. In contrast, the modular architecture contains two recurrent modules. Task identity determines how inputs are routed: task A stimuli are delivered to module \(M_A\), and task B stimuli are delivered to module \(M_B\). The outputs of the modules are combined by a shared readout. Let \(x_t\) denote the input at time \(t\), \(h_t^A\) and \(h_t^B\) the hidden states of the two modules, and \(y_t\) the output. In the modular model, the input is partitioned into task-specific slices, $x_t^A = m_A \odot x_t$, $x_t^B = m_B \odot x_t$, where \(m_A\) and \(m_B\) are binary masks selecting the task-relevant input dimensions. Each module evolves according to
        \[
        h_t^A = \tanh\!\big(W_{\mathrm{ih}}^A x_t^A + W_{\mathrm{hh}}^A h_{t-1}^A\big),
        \]
        \[
        h_t^B = \tanh\!\big(W_{\mathrm{ih}}^B x_t^B + W_{\mathrm{hh}}^B h_{t-1}^B\big),
        \]
        with recurrent processing occurring separately in each module and no inter-module recurrent communication in the main analysis. The pre-readout state is the concatenation of the two module states
        and the shared readout maps this state to four outputs (two cosine--sine pairs):
        \[
        y_t = W_{\mathrm{out}} h_t.
        \]
        
        For the single-network baseline, the same recurrent update is applied to a single hidden population:        $h_t = \tanh\!\big(W_{\mathrm{ih}} x_t + W_{\mathrm{hh}} h_{t-1}\big)$, $y_t = W_{\mathrm{out}} h_t$. This design isolates the effect of task-partitioned input processing and recurrent separation while keeping the output format comparable across architectures.
        
        \subsection{Training Procedure}
        
        Each model was initialized from a new random seed and trained on a single task sequence following the $A1 \rightarrow B \rightarrow A2$ protocol. Different model runs, therefore, correspond to different trial sequences derived from the original experimental schedules with independent random initializations. To vary the representational regime, we rescaled all trainable parameters after default PyTorch initialization by a global factor $\gamma$ (Figure~\ref{fig:overview}c). Larger values of $\gamma$ increase the magnitude of the initial weights, whereas smaller values reduce it. This manipulation is commonly associated with transitions between so-called lazy and rich learning regimes, which in turn are linked to differences in effective representational dimensionality \cite{flesch2021rich}. In this work, we treat $\gamma$ as a practical probe of representational regime, without assuming that it provides a pure or isolated control of dimensionality, as it may also affect optimization dynamics and conditioning. All models were trained using MSE loss on a cosine-sine encoding of the target angle. On each trial, the loss was computed only on the output components corresponding to the currently probed feature, ensuring that learning signals matched the task structure of the sequential protocol. 
        
        \subsection{Behavioral Measures}

        We quantified model performance using accuracy, transfer, and interference. Accuracy refers to angular prediction accuracy after reconstruction from the cosine-sine output. Transfer was measured as mean winter-trial accuracy in the first six trials of phase \(B\) minus mean winter-trial accuracy in the last six trials of phase \(A1\), capturing how much prior learning facilitates early performance on the new task. Interference was quantified from responses in phase \(A2\) by fitting a von Mises mixture to $A2$ winter responses and taking one minus the mixture weight on the task-A component. This measures how much behavior on task A has shifted toward task B after learning task B, and thus reflects the degree to which previously learned responses have been overwritten.
        
        \subsection{Representational Analysis}
        
        To characterize the learned hidden-state geometry, we extracted the hidden representation at the final time step for each stimulus and phase. For each phase, the hidden states were stacked into a data matrix and analyzed using principal component analysis (PCA). We used the number of principal components required to explain 99 \% of the variance as a measure of effective dimensionality. We also computed principal angles between task-specific subspaces. Using hidden states after phase \(B\) from a canonical stimulus sweep, we split the rows into two groups according to the feature associated with each stimulus under the stored ordering and fitted a PCA within each group. We then reported the largest principal angle between the resulting two-dimensional subspaces. For the qualitative geometry analysis, we projected hidden states onto the first three principal components of a single PCA fit computed jointly over the \(A1\), \(B\), and \(A2\) phases. These 3D PCA projections provide an intuitive illustration of how task representations reorganize across learning and how this reorganization depends on architecture and initial weight scale.
    
\section{Results}

\begin{figure*}[t]
            \centering
            \includegraphics[width=\linewidth]{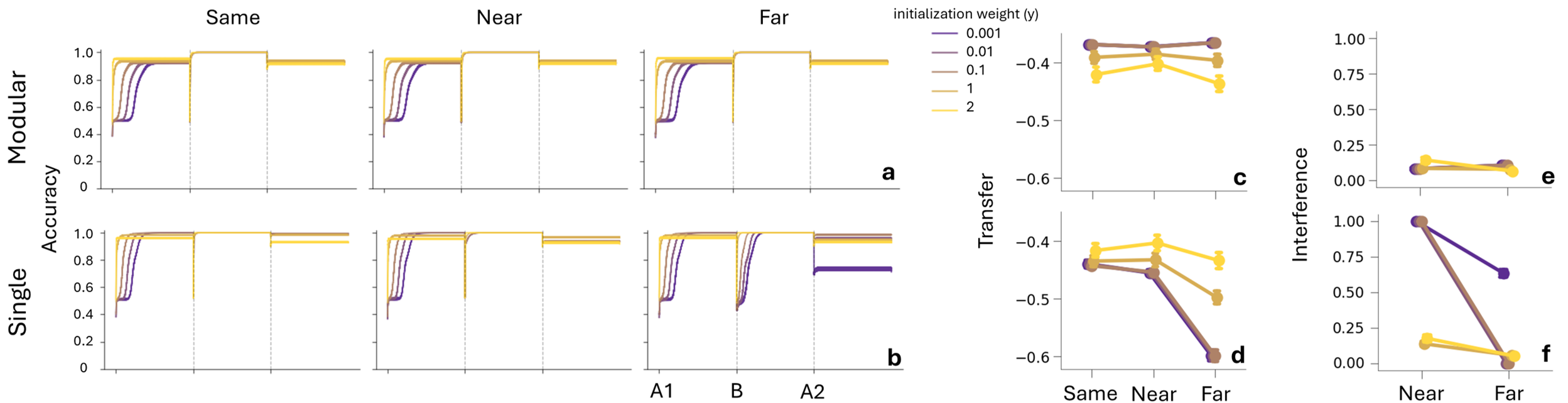}
            \caption{\textbf{Modular structure attenuates transfer-interference costs in sequential learning in constrained regimes}. \textbf{(a, b)} Accuracy across the sequential $A1 \rightarrow B \rightarrow A2$ training protocol for the modular network \textbf{(a)} and the single network \textbf{(b)} under the three task-similarity conditions (same, near, far) and across initialization weight scales. In both architectures, learning on $A1$ and $B$ rapidly reaches high accuracy, but the single network shows a stronger drop in $A2$ performance in the far condition at the smallest initialization scales. \textbf{(c, d)} Transfer metric as a function of task similarity and initialization scale for the modular \textbf{(c)} and single \textbf{(d)} architectures. The modular network shows relatively stable low transfer across similarity conditions and weight scales, whereas the single network becomes more sensitive to task dissimilarity, with the far condition showing the strongest degradation for small $\gamma$. \textbf{(e, f)} Interference metric for the modular \textbf{(e)} and single \textbf{(f)} architectures. Interference remains low in the modular network across conditions, while the single network shows substantially larger interference, especially in the near condition and at the lowest initialization scales. The full figure shows that architectural separation reduces susceptibility to sequential interference, but its benefit is most visible when representational constraints are strong.}
            \label{fig:accuperf}
        \end{figure*}

        Figure \ref{fig:accuperf} shows the behavioral consequences of the modulation of $\gamma$ as a proxy for the representational regime.
        
        In the modular network, accuracy remains high across the training sequence for all similarity conditions and $\gamma$, indicating robustness to sequential task exposure. Transfer changes only modestly across conditions, and interference remains low throughout, confirming the intuition that modular input-separation buffers the system against overwrite during task switching. The single network is sensitive to task similarity and changing $\gamma$. While it also learns the first two phases well, $A2$ performance drops more visibly in the far condition at the lowest $\gamma$ values, indicating a stronger transfer–interference cost as the task relation becomes less alike. 
        
\begin{figure}[!t]
            \centering
            \includegraphics[width=3.4in]{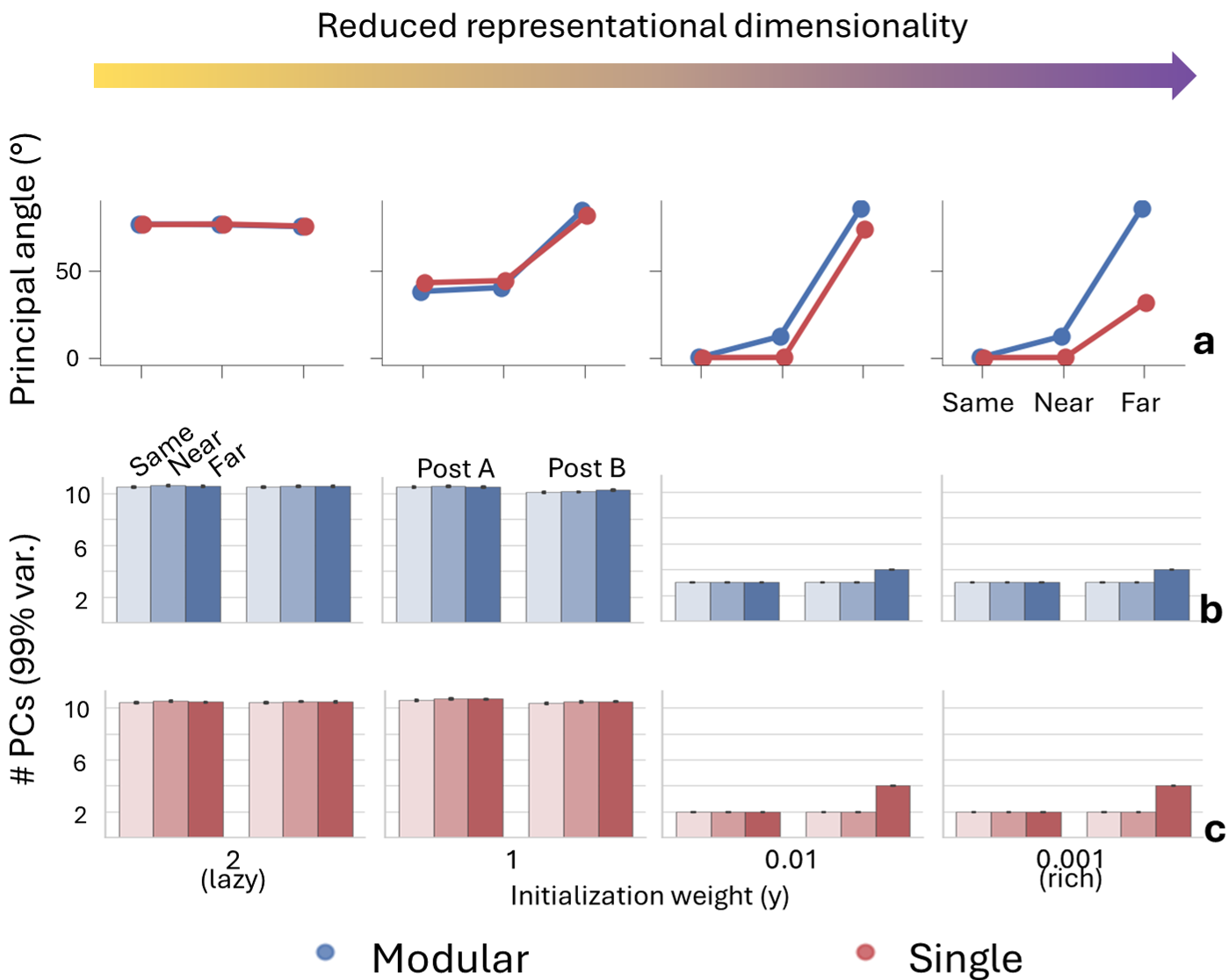}
            \caption{\textbf{Initialization weight scaling controlled representational dimensionality reveals architecture-dependent representational geometry.} \textbf{(a, b)} Effective dimensionality of hidden representations, measured as the number of principal components required to explain 99\% of the variance, for modular (\textbf{a}) and the single networks (\textbf{b}). Columns correspond to decreasing initialization weight scale, from the lazy regime to the richer, lower-scale regime. Within each column, bars show the task-similarity conditions. Dimensionality remains high at large initialization scales and decreases strongly at smaller scales. \textbf{(c)} Principal angles between task-specific representational subspaces for the same, near, and far conditions across the same initialization scales. In the high-dimensional regime, both architectures show similar geometry. In the lower-dimensional regime, the modular network exhibits a clearer similarity-dependent structure, with stronger separation for far tasks and intermediate alignment for near tasks. Taken together architectural separation becomes meaningful only once representational dimensionality is sufficiently reduced, at which point task similarity begins to shape subspace organization.}
            \label{fig:reduceddim}
        \end{figure}

        The transfer and interference summaries reinforce this pattern: compared with the modular network, the single network shows a less stable transfer profile and larger interference effects with decreasing initialization weight values. 
        Figure \ref{fig:reduceddim} shows that the effect of architectural separation strongly depends on the representational regime. At large $\gamma$ values, corresponding to the lazy regime, both the modular and single architectures maintain high effective dimensionality and display broadly similar subspace geometry across task similarity conditions. Here, the architecture has little influence on the structure of the learned representations.
        
        As $\gamma$ decreases, the effective dimensionality decreases dramatically. This reduction is accompanied by a stronger dependence of representational geometry on task similarity. In the modular network, the same condition remains closely aligned, the near condition occupies an intermediate position, and the far condition is pushed toward stronger separation. The single network shows a weaker and less structured dependence on task similarity, indicating less controlled subspace organization. The figure suggests that here modularity becomes effective only when the representation is sufficiently constrained by lower representational dimensionality. In that regime, architecture begins to shape how task representations are allocated in the hidden space, and task similarity determines whether they overlap, partially align, or separate more strongly. The results indicate that structural separation becomes geometrically effective when the allocation of subspace structure is nontrivial.

\begin{figure*}[!t]
            \centering
            \includegraphics[width=\linewidth]{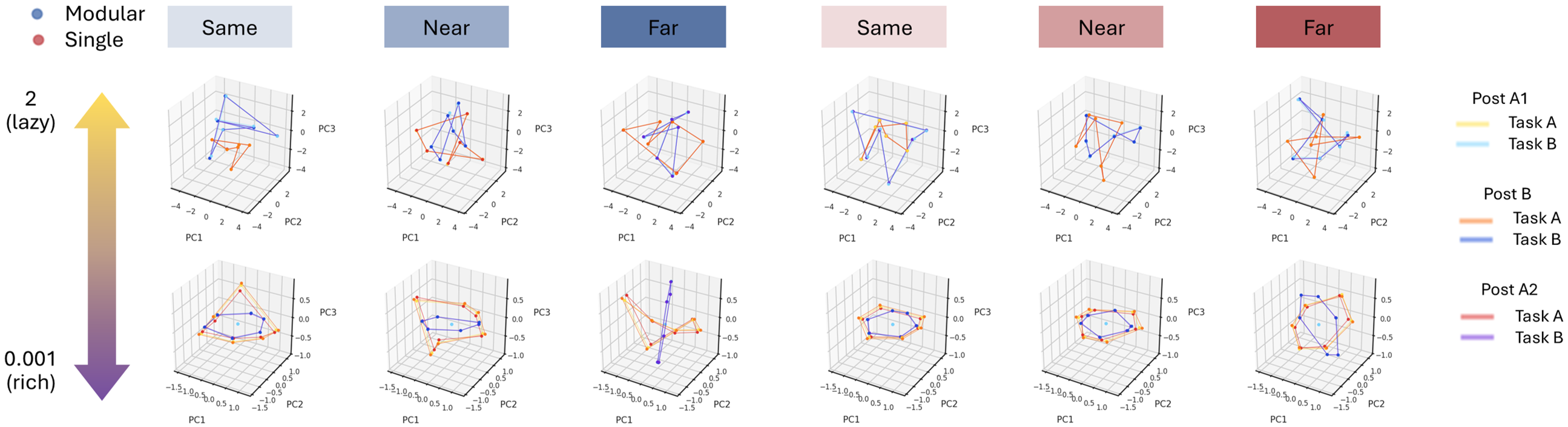}
            \caption{\textbf{3D PCA projections reveal similarity-dependent organization of task representations under reduced dimensionality.} Each panel shows hidden-state trajectories projected onto the first three principal components of a PCA fitted jointly to Post $A1$, Post $B$, and Post $A2$ activations for a given network and initialization regime. For each phase, trajectories are computed over a fixed sweep of 12 inputs, with the first six corresponding to Task A and the last six to Task B. For each phase and task, closed loops connect stimuli in sweep order. The left block shows the modular network, and the right block shows the single network. Columns correspond to task similarity conditions (same, near, far). The top row shows the lazy (high-dimensional) regime, and the bottom row shows the rich (low-dimensional) regime. Colors encode phase and task identity (Task A: warm colors, Task B: cool colors). In the high-dimensional regime (top row), trajectories occupy a broad state space and exhibit relatively weak structure, with limited differences between architectures. In the low-dimensional regime (bottom row), trajectories become more compact and their relative arrangement reflects task similarity more clearly. In the modular network, same-task trajectories remain relatively closely aligned across phases, near tasks exhibit intermediate reorganization, and far tasks show stronger separation. This similarity-dependent structure is less pronounced in the single network, which in turn exhibits closer alignment in the same regime. Overall, the figure illustrates that reduced representational dimensionality reveals a constrained geometric organization of task representations, within which architectural separation leads to more structured, similarity-dependent alignment.}
            \label{fig:reprgeom}
        \end{figure*}

        Figure \ref{fig:reprgeom} provides a qualitative view of the representational geometry underlying the quantitative results in Figures \ref{fig:accuperf} and \ref{fig:reduceddim}. In the lazy regime (y=2), hidden-state trajectories occupy a relatively broad state space, and the geometry is broadly similar across the modular and single architectures. Although the different task-similarity conditions are visible, the structure is not strongly constrained, consistent with the earlier finding that architectural effects are weak when the representation remains high-dimensional.
        In the rich regime (y=0.001), the trajectories collapse into a more compact, low-dimensional geometry. Here, the effect of task similarity becomes apparent, especially in the modular network: same-task trajectories remain more closely aligned across phases, near-task trajectories show intermediate reorganization. Far-task trajectories exhibit stronger separation. 
        
        The single network also shows a reduction in dimensionality, but the resulting geometry is less clearly adapted to task similarity. The PCA projections support the interpretation that representational dimensionality gates the extent to which architectural separation can shape task geometry. When dimensionality is high, both architectures can represent the task sequence with comparatively little geometric constraint. When dimensionality is reduced, the modular architecture yields a more structured and similarity-sensitive organization of hidden states.
        To assess the robustness of these findings, we conducted ablations varying module size and input routing. The qualitative patterns reported above remained unchanged. Results for all ablations are provided in the code repository.

\section{Discussion}

        Our results show that the effect of architectural separation in continual learning is conditional rather than universal. Across the sequential $A1 \rightarrow B \rightarrow A2$ protocol, the modular network generally preserved performance better and exhibited lower interference than the single network, but this advantage was not present uniformly across all regimes. The key factor was the representational regime induced by $\gamma$. In the high-dimensional, lazy regime, both architectures behaved similarly, whereas in the lower-dimensional, richer regime, differences in transfer, interference, and representational geometry became more pronounced. This suggests that architectural separation is not intrinsically beneficial in all settings, but becomes effective when the representational space is sufficiently constrained for structural bias to matter. 
        When dimensionality is high, the network has enough degrees of freedom to encode successive tasks with relatively little geometric pressure, so the influence of architectural separation remains weak. When dimensionality is reduced, however, the allocation of representational subspace becomes a binding constraint. In this regime, task representations must compete for a more limited set of directions, and architectural structure begins to shape how those directions are used. The results therefore support a view in which representational dimensionality gates the functional impact of architecture. This interpretation is reinforced by the geometric analyses. The principal-angle measures show that in the low-dimensional regime, the modular network exhibits a more structured dependence on task similarity: the same condition remains closely aligned, the near condition occupies an intermediate position, and the far condition separates more strongly. The single network shows a weaker pattern. The 3D PCA projections provide a qualitative illustration of the same effect. In the lazy regime, trajectories occupy a broader space, and the difference between architectures is limited. In the rich regime, trajectories become more compact, and their arrangement reflects the gradient of task similarity more clearly.
        
        These analyses indicate that modularity does not simply reduce interference in an absolute sense; it supports a graded organization of representational geometry when the network operates under capacity constraints. An important aspect of the present architecture is that separation is structural but not total. The modular network routes task-specific inputs into different recurrent modules, but the two modules still share a common readout. This matters as it means the model is not learning two completely independent task systems. Instead, it must maintain some degree of coordination at the output level, even when the recurrent pathways are separated. This shared readout may help explain why the geometry we observe is graded rather than fully discrete: the network can preserve task-dependent separation internally while still being coupled through a common behavioral objective. Thus, the modular architecture supports a constrained form of division of labor rather than complete isolation. 
        
        These findings have two implications for continual learning. First, the relevant question is not simply whether a model is modular or not, but when modularity becomes functionally meaningful. Second, the goal of continual learning should not be interpreted as maximal separation between tasks. A more appropriate target is similarity-dependent geometry: overlapping representations when tasks are effectively the same, partial reorganization when tasks are related, and stronger separation when tasks are dissimilar. From this perspective, continual learning is a problem of adaptive representational allocation rather than a binary choice between sharing and isolation.
        
        Several limitations should be acknowledged. First, effective dimensionality is measured here using PCA-based summaries, which are informative but remain approximate proxies for the true intrinsic dimension of the representation. Second, the initialization scale should be interpreted with care, as it likely affects not only dimensionality but also optimization dynamics, conditioning, and how easily the architectural biases can be expressed. Third, the 3D PCA plots are illustrative rather than definitive: they support the quantitative analyses but do not replace them. Finally, this study focuses on a restricted A-B-A sequential structure. This design is appropriate for reproducing a specific transfer-interference pattern from human studies, but it does not capture longer-term adaptation across extended task sequences.

        Future work could extend this framework in several directions. Longer task sequences would allow us to examine how architectural bias and representational geometry evolve over repeated episodes of interference and recovery. It would also be informative to study sequences in which task similarity changes over time, for example, by moving from same to near to far conditions within a single learning trajectory, to test how previously established representations adapt when later tasks become less or more similar. Another promising direction is to examine whether dimensionality itself can be regulated dynamically during learning, rather than being set indirectly through initialization scale. Finally, the present architecture represents a specific form of task-dependent input partitioning and recurrent separation; future work could compare this design to alternative modular architectures, including versions with inter-module communication, to determine how different kinds of structural coupling alter geometry and transfer-interference behavior.
        
        Overall, the main contribution of this work is to show that modularity is a mechanism whose effect depends on the representational regime. When the network is sufficiently constrained, architectural separation can shape task geometry in a similarity-dependent manner and reduce transfer-interference costs. This suggests that future continual learning systems may benefit less from fixed notions of modularity than from mechanisms that regulate representational dimensionality and subspace allocation dynamically as task structure changes.

\ifanonymous\else
\section{Acknowledgements}
Funded by the European Union (ERC, GROW-AI, 101045094). Views and opinions expressed are however those of the authors only and do not necessarily reflect those of the European Union or the European Research Council. 
\fi

\footnotesize
\bibliographystyle{apalike}
\bibliography{example} 

\end{document}